%% file: 6920.HeJ.tex
\documentclass[letterpaper]{article} 
\usepackage{aaai23}  
\usepackage{comment}
\usepackage{times}  
\usepackage{helvet}  
\usepackage{courier}  
\usepackage[hyphens]{url}  
\usepackage{graphicx} 
\usepackage{natbib}  
\usepackage{caption} 
\usepackage[ruled, linesnumbered]{algorithm2e}
\usepackage{enumerate}

\input{commands}

\title{MSDC: Exploiting Multi-State Power Consumption in Non-intrusive Load Monitoring based on A Dual-CNN Model}
\author{
    Jialing He, \textsuperscript{\rm 1,3}
    Jiamou Liu, \textsuperscript{\rm 2}
    Zijian Zhang,\textsuperscript{\rm 3,4 \thanks{Corresponding author.}}
    Yang Chen, \textsuperscript{\rm 5}
    Yiwei Liu, \textsuperscript{\rm 6}
    Bakh Khoussainov, \textsuperscript{\rm 7}
    Liehuang Zhu, \textsuperscript{\rm 3}
}
\affiliations{
    \textsuperscript{\rm 1} College of Computer Science, Chongqing University, Chongqing, China, 400044.\\
    \textsuperscript{\rm 2} School of Computer Science, The University of Auckland, Auckland 1142, New Zealand.\\
    \textsuperscript{\rm 3} School of Cyberspace Science and Technology, Beijing Institute of Technology, Beijing, China, 100081.\\
    \textsuperscript{\rm 4} Southeast Institute of Information Technology, Beijing Institute of Technology, Fujian China, 351100.\\
    \textsuperscript{\rm 5} Strong AI Lab, The University of Auckland, Auckland 1142, New Zealand.\\
    \textsuperscript{\rm 6} Defence Industry Secrecy Examination and Certification Center, Beijing, China, 100089.\\
    \textsuperscript{\rm 7} School of Computer Science and Engineering, University of Electronic Science and Technology of China, Chengdu, China, 611731.\\
    hejialing@cqu.edu.cn, \{jiamou.liu, yang.chen\}@auckland.ac.nz, \{zhangzijian, liehuangz\}@bit.edu.cn, yiweiliu\_disecc@163.com, bmk@uestc.edu.cn
}

\begin{document}

\maketitle

\begin{abstract}
Non-intrusive load monitoring (NILM) aims to decompose aggregated electrical usage signal into appliance-specific power consumption and it amounts to a classical example of blind source separation tasks. Leveraging recent progress on deep learning techniques, we design a new neural NILM model {\em Multi-State Dual CNN} (MSDC). Different from previous models, MSDC explicitly extracts information about the appliance's multiple states and state transitions, which in turn regulates the prediction of signals for appliances. More specifically, we employ a dual-CNN architecture: one CNN for outputting state distributions and the other for predicting the power of each state. A new technique is invented that utilizes conditional random fields (CRF) to capture state transitions. Experiments on two real-world datasets REDD and UK-DALE demonstrate that our model significantly outperform state-of-the-art models  while having good generalization capacity, achieving 6\%-10\% MAE gain and 33\%-51\% SAE gain to unseen appliances.

\end{abstract}

\section{Introduction}
\label{intro}
Energy efficiency amounts to one of the major challenges facing today's families \cite{AlahakoonY16}.
Smart grids, through their ability to meter and monitor energy consumption to the level of individual households, provide fine-grained data to discover patterns in a user's behaviors. With these patterns, appropriate strategies can be implemented to improve power efficiency \cite{wilson2015smart}. A critical issue of this technology is {\em non-intrusive load monitoring} (NILM): dissecting aggregated energy consumption signals of a household -- the usual input to smart meters -- into per-appliance signals \cite{hart1992nonintrusive}. NILM is important not only to accurately unravel the household's power consumption patterns \cite{ccimen2020microgrid,hassan2022differentially}, but also to tasks such as load forecasting  \cite{dinesh2019residential, wang2018learning, kong2017short} and malfunction detection \cite{shao2017novel,green2019dashboard,rashid2019can}.

NILM is a classical example of a single-channel {\em blind source separation} (BSS) task such as the infamous {\em cocktail party problem}, i.e., extracting multiple data sources from a single mixed observation. BSS is known to be highly underdetermined \cite{naik2014blind}. To date, the mainstream solutions for BSS have relied on statistical modeling. In the NILM literature, hidden Markov models (HMM) \cite{zia2011hidden} and conditional random fields (CRF) \cite{azaza2017finite} are two popular modeling paradigms to capture the operations of household appliances. As an appliance's power consumption signal is very much dictated by (1) the appliances' states which have (stable) power consumption levels, i.e., {\em power states}, and (2) the patterns of the power state transitions, these models embed appliances' power states and state transitions. This information is in turn  used for statistical inference of power signals. A severe limitation of these mainstream solutions is computational complexity. For large households that contain many appliances with potentially many power states, these methods tend to incur unreasonable computational costs. Moreover, they perform poorly for unseen appliances as the inferred model only fits appliances used for training.

Recent advances in deep learning techniques have presented new pathways to address these limitations. More specifically, one can model the NILM problem as a {\em sequence-to-sequence} (seq2seq) learning task: Training a deep learning model that maps the aggregated power signal to the target per-appliance power signals. Based on this idea, several NILM solutions have been introduced utilizing various neural network architectures such as convolutional neural networks (CNNs), recurrent neural networks (RNNs), long-short-term memory (LSTM), denoising autoencoders \cite{kelly2015neural,mauch2015new,hsu2019self}, and transformers \cite{yue2020bert4nilm}.

In this paper, we argue that the existing neural-based models are insufficient for NILM. In particular, these models were mostly designed to directly translate input aggregated signals into per-appliance signals, without the necessary step of exploiting power states and state transitions of the appliances. As demonstrated in classical statistical models, power state and transitions of appliances can provide key knowledge on the power consumption of appliances and could greatly strengthen the predictive ability of a model. Further, uncovering these power states and patterns of state shifts may endow the prediction results with better interpretability. Our goal is thus to develop a neural-based NILM model through mining power states and state transitions.

Our main contributions are three fold:
{\bf(1)} We define a formal model of the {\em multi-state NILM problem} and provide a theoretical justification (see Theorem~\ref{thm:power}, Sec.~\ref{sec:problem}) for the advantage of the multi-state setting on variance reduction in sampling power data.
{\bf(2)} Accordingly, we propose a novel multi-state NILM model (called MSDC, see Sec.~\ref{sec:model}) that features a dual-CNN architecture: one CNN ({\em{state-CNN}}) for capturing the appliance's multiple power states, the other ({\em{value-CNN}}) for predicting the power consumption for each state.  A cross-entropy is added to {\em{state-CNN}} as a regularization item, which degrades the error in predicting the appliance's power consumption.  We further replace the cross-entropy loss with a CRF \cite{LaffertyMP01} regularization that enables capturing state transitions.
{\bf(3)} Experimental results on two real-world datasets REDD \cite{kolter2011redd} and UK-DALE \cite{kelly2015uk} (see Sec.~\ref{sec:experiments}) show that our model generalizes well to unseen appliances where it achieves $6\%$-$10\%$ MAE gain and $33\%$-$51\%$ SAE gain over the state-of-art on the accuracy of energy decomposing.
    Moreover, visualization results demonstrate that the power state transitions predicted by our model align with the ground truth at a higher level, compared with baselines.

\section{Related Work}\label{sec:related}
NILM was first proposed in \cite{hart1992nonintrusive}. Mainstream solutions of the problem employ statistical inference models such as HMM and CRF. \cite{zia2011hidden} modelled an aggregated signal as a combination of HMMs, each corresponding to an appliance, and found that power consumption patterns of appliances can be differentiated from the aggregated profile.  \cite{kim2011unsupervised} investigated several variants of HMM and demonstrated that a conditional factorial HMM integrating additional features about the time usage of appliances outperforms others. Subsequent HMM variants further demonstrated non-trivial performance gain \cite{kolter2012approximate,kong2016hierarchical,mauch2016novel}. Nevertheless, the assumption that any observation is independent of the other may violate real-life situations and lead to label bias problem.

CRF-based methods relax the independent assumption  for observations and utilize the contextual information from all observations to mitigate the label bias problem.  \cite{azaza2017finite} exploited CRF and the clustering algorithm to capture the component appliance's power signal and corresponding on-off states. This method was extended by \cite{he2019novel} to capture appliances' multiple states. However, high computational complexity and poor scalability prevent these statistical modeling methods from practical usage. Our proposed model avoids these issues because it does not explicitly maintain a state transition network, but rather capture transitions implicitly using CNNs.

Neural networks are suitable for NILM thanks to their high expressibility.  \cite{kelly2015neural} first tackled NILM using deep neural networks such as CNN, LSTM, and denoising autoencoders, with superior performance over statistical modeling methods. A series of subsequent CNN-based models \cite{zhang2018sequence,shin2019subtask,chen2019scale} revealed that CNN provides a versatile framework for extracting latent features such as power thresholds, change point and duration for the appliance's power consumption, thereby outperforming models based on other neural networks.
As CNN-based models are shown to exhibit consistently reliable performance, we also adopt CNN in this work.

Other models leveraged the correlations between whether an appliance is on or off and its power \cite{shin2019subtask}. A sub-network is added to capture appliances' on-off states, which expedite identifying power signals. Several models have been proposed using dilated convolutions \cite{chen2019scale}, generative adversarial network \cite{pan2020sequence}, and attention mechanisms \cite{sudoso2019non,piccialli2021improving}. However, these models are limited to predicting whether the appliance is on or off, without making an effort to reveal the full spectrum of power states. In some sense, one can view capturing the on-off states of an appliance as an extremely coarse approximation of the power states. Yet, doing this means that all different power levels when the appliance is turned on are regarded as being in one ``on'' state, hence does not provide enough information to deduce the power signal.

\begin{figure}
	\centering
    \includegraphics[width=.25\textwidth]{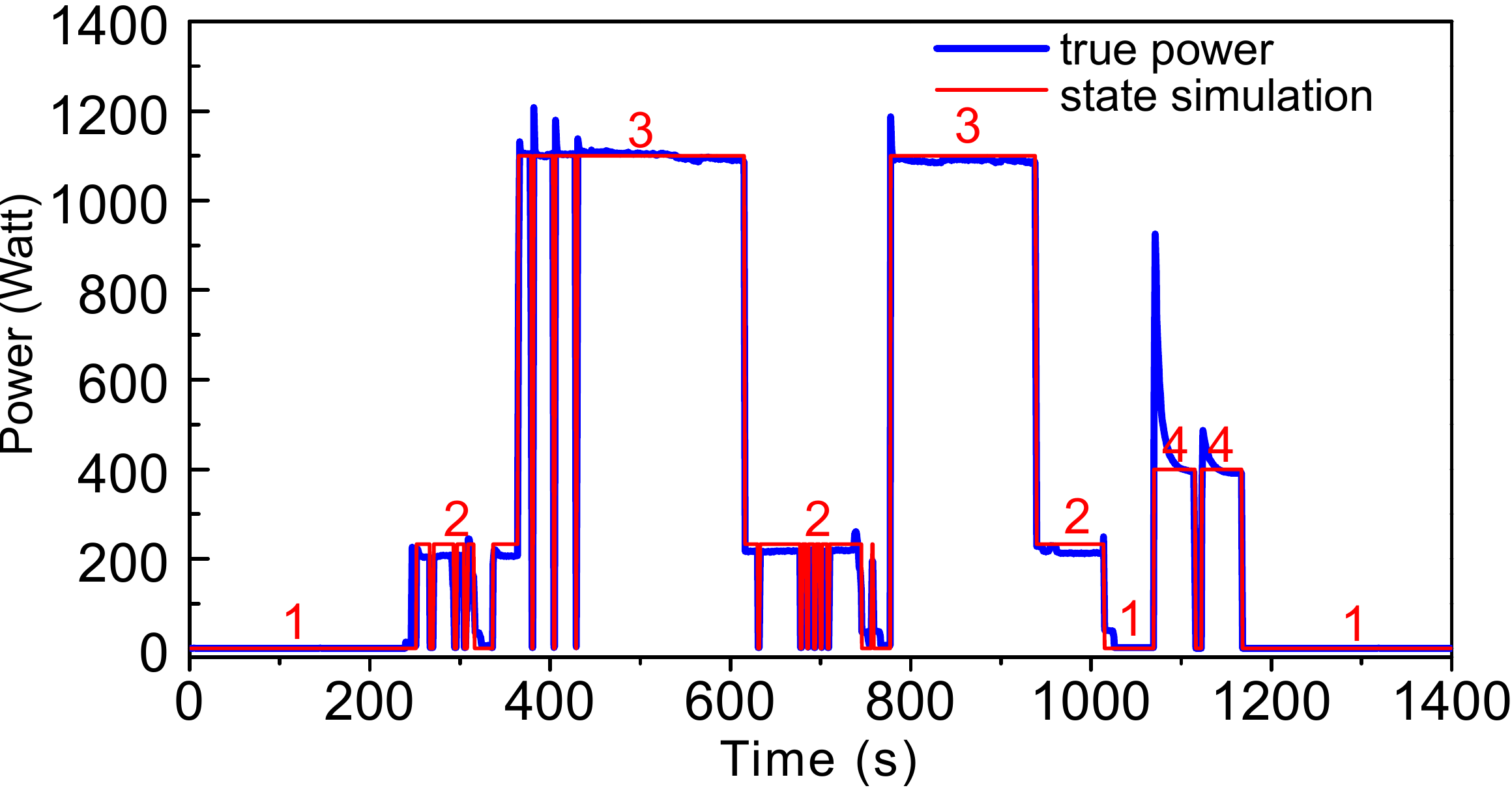}
    \vspace{-3mm}
    \caption{Multi-state power consumption of a dishwasher in REDD dataset. Blue and red curves represent truth power signals and simulated power states, respectively.}
    \label{fig:truth}
\vspace{-10pt}
\end{figure}
\section{Problem Formulation}\label{sec:problem}
{\em Multi-state non-intrusive load monitoring} (MS-NILM) problem seeks to recover the power consumption of individual multi-state appliances 
from the aggregated power signal. Let $\Xv =\left(x_1, x_2, \ldots, x_T\right) \in \mathbb{R}_+^T$ denote the {\em aggregated power}, where $T \in \mathbb{N}_+$ represents the {\em measured time}. Suppose the scope of investigation includes $N$ appliances. 
Appliance $i$'s power consumption is denoted by $\Yv^i=(y_1^i, y_2^i, \ldots , y_T^i) \in \mathbb{R}_+^T$. Set $[N]$ as $\{1,2,\ldots, N\}$.
Following standard assumption in NILM literature, we assume  $\forall i \in [N], t \in [T]$,
the signal $x_t=\sum_{i=1}^N y_t^i+z_t+\epsilon_t$,
where $z_t$ denotes the total power consumption of appliances included in the power reading but fall {\em outside} of our scope of investigation, and  $\epsilon_t$ is a noise signal \cite{zhang2018sequence}.

A multi-state appliance can be modeled as a {\em finite state machine} with a fixed set of {\em operational states}. As we are interested only in power consumption, we view two {\em operational states} as in the same {\em power state} if they have the same power consumption level, and transitions between operational states are grouped into transitions between power states. For appliance $i$, let $M^i\in \Nbb_+$ denote the number of its power states.  In principle, each power state is characterized by a unique power level, and at each time step the appliance will be in one power state $s_t^i$. In reality, this assumption may be over-simplified: (1) The actual power reading for the appliance at a state is prone to perturbation so there could be small fluctuation at the fixed power level. Therefore it makes sense to individually express the power consumption level $c_{t,j}^i$ for appliance $i$ at state $j$ and at given time $t$. (2) Due to uncertainty within the learning process, it is difficult to pinpoint a single state for the appliance to be operating in at a given time, but rather a probability $p_{t,j}^i$ for appliance $i$ to be in state $j$ at time $t$.
%
More formally, the power consumption signal for states at time step $t$ is determined by power consumption at all $M^i$ states $\Cmc_t^i=(c_{t,1}^i, c_{t,2}^i, \ldots, c_{t,M^i}^i) \in \Rbb_+^{M^i}$ and probability distribution $\Pmc_t^i = ( p_{t,1}^i, p_{t,2}^i, \ldots, p_{t,M^i}^i ) \in [0,1]^{M^i}$, where $c_{t,j}^i \neq c_{t,k}^i$ for $j \neq k \in [M^i]$ 
and $p_{t,s}^i$ is the probability $\Pr(s_t^i = s)$. Notably, the power state $s_t^i$ generally depends on previous states $s_1^i,\ldots, s_{t-1}^i$. Indeed, the state transitions would follow certain specific patterns. E.g., in Fig.~\ref{fig:truth} State 3 is a successor of State 2. Following standard convention, we assume that the state transition is Markovian, i.e., $\Pr(s_t^i \vert s_{t-1}^i,\ldots, s_1^i) = \Pr(s_t^i \vert s_{t-1}^i)$. 
Given above settings, an appliance's power consumption can be viewed as the expectation on that of each state:
\begin{equation}\label{eq:appliance-power}
    y_t^i=\Cmc_t^i \cdot \Pmc_t^i, ~~ \forall i \in [N], t \in [T].
\end{equation}
MS-NILM can thus be addressed by inferring each appliance's power state sequence and the corresponding power consumption from the aggregated power signals; at each time step, each appliance's power consumption is subsequently estimated according to Eq.~\eqref{eq:appliance-power}. 
To summarize, we state the problem of MS-NILM as follows:

\begin{description}
    \item[Problem Statement:] {\em Given a set of instances of energy disaggregation of the form $\{(\Xv, \Yv^1,\ldots, \Yv^N)\}$, infer $\Cmc_t^i$ and $\Pmc_t^i$ for all $i\in [N]$ and  $t \in [T]$, and further generalize to unseen aggregated power signal $\Xv'$.}
\end{description}

\subsection{Justifications for the Multi-state Setting}
\label{sec:theory}
We next discuss the need for a multi-state setting for NILM. The idea is to show that sampling power data from a fine-grained state structure can result in a smaller sample variance, thus attaining a higher probability to reach the mean of real power data. We next make some useful assumptions, based on which we present our main theoretical result. Following conventional practice in NILM literature, we assume that an appliance's power consumption follows a normal distribution\cite{zhang2018sequence,mauch2016novel}.

\begin{assumption}\label{ass:normal}
	For each appliance $i$, the power consumption of each state $s$ at each time step $t$ is drawn from a normal distribution, i.e., $c_{t,s}^i \sim \Nmc(\mu_{s}^i, \sigma_{s}^i)$.
\end{assumption}
Recall that the observed power is the expectation on the power of all states, i.e., $y_t^i = \sum_{s=1}^{M^i} p_{t,s}^i c_{t,s}^i$. Due to the independence of power measures of all states and the additivity of normal distributions, we have the following fact.
\begin{fact}\label{fact:normal-dist}
Under Assumption~\ref{ass:normal}, the input power $y_t^i$ of an appliance also follows a normal distribution $\Nmc(\mu^i, \sigma^i)$ such that
	$\mu^i = \sum_{s=1}^{M^i} p_{t,s}^i \mu_s^i$ and $(\sigma^i)^2 = \sum_{s=1}^{M^i} (p_{t,s}^i \sigma^i_s)^2$.
\end{fact}
Since a multi-state model decomposes the total power into a fine-grained state structure, it is natural to propose the following assumption that enforces the variance of the power of each state to not exceed that of the observed total power. This can be seen from the truth depicted in Fig.1: a dishwasher has a steady power level at each state; while, the variance would be increased if we merge any two states into an abstract state.

\begin{assumption}\label{ass:var}
For all $s\in [M^i]$, $\sigma_s^i \leq \sigma^i$.
\end{assumption}

We are now ready to present our main result that uses the following notations: $\tilde{y}_t^i$ denotes the sampled power under the single-state assumption; $\bar{y}_t^i = \Ebb_{s\in [M^i]} [\bar{c}_{t,s}^i]$ denotes the sampled power under the multi-state setting which is obtain by averaging the sampled power $\bar{c}_{t,s}^i$ of each state $s \in [M^i]$ (suppose the state information is known a priori).
We show in the following theorem that the sampled power data can enjoy a reduced sample variance from the multi-state setting.

\begin{theorem}\label{thm:power}
Suppose we have a sufficiently large number of independent samples. Under Assumptions~\ref{ass:normal} and \ref{ass:var},
the expectations and variances of $\tilde{y}_t^i$ and $\bar{y}_t^i$ satisfy $\mathbb{E}[\bar{y}_t^i]=\mathbb{E}[\tilde{y}_t^i]$ and $\mathbb{D}[\bar{y}_t^i] \leq \mathbb{D}[\tilde{y}_t^i]$ for all $t \in [T]$, where the inequality is strict if $M^i \geq 2$.
\end{theorem}

\begin{proof}
With a sufficiently large number of independent samples, both $\tilde{y}_t^i$ and $\bar{y}_t^i$ would approach normal distributions, i.e., $\tilde{y}_t^i \sim \Nmc(\mu^i, \sigma^i)$ and $\tilde{y}_t^i = \sum_{s=1}^{M^i} p_{t,s}^i c_{t,s}^i$.
Then, according to Fact.~\ref{fact:normal-dist}, $\mathbb{E}[\hat{y}_t^i]=\mathbb{E}[\tilde{y}_t^i]$ follows immediately from the additivity of normal distributions. The reduced variance can be derived from Assumption~2 as follows:
\begin{equation*}
\small
\begin{aligned}
	\mathbb{D}[\hat{y}_t^i]& = \mathbb{D}\left[\sum\nolimits_{s=1}^{M^i} p_{t,s}^i c_{t,s}^i\right]=\sum\nolimits_{s=1}^{M^i} p_{t,s}^2 \mathbb{D}[c_{t,s}^i]\\
	& =\sum\nolimits_{s=1}^{M^i} p_{t,s}^2 \sigma_s^i \leq \sum\nolimits_{s=1}^{M^i} p_{t,s}^2 \sigma^i \leq \left(\sum\nolimits_{s=1}^{M^i} p_{t,s}\right)^2 \sigma^i \\
	& =\sigma^i=\mathbb{D}[\tilde{y}_t^i].
\end{aligned}
\end{equation*}
If $M^i \geq 2$, we have $\sum_{s=1}^{M^i} p_{t,s}^2 < \left(\sum_{s=1}^{M^i} p_{t,s}\right)^2$ and hence the inequalities above is strict.
\end{proof}
\begin{corollary}
When the appliance has $M^i (M^i\geq2)$ power states in total, then the power estimation regarding the appliance in a multi-state setting with $M^i$ states would approach the mean value of truth power data with a higher probability than using the single-state setting.
\end{corollary}
\vspace*{-0.2cm}
\begin{proof}
Assume $\bar{y}_t^i=\sum_{s=1}^{M^i} p_{t,s}^i c_{t,s}^i$  and $\tilde{y}_t^i$ represent the power estimation utilizing $M^i$ states and one state respectively at each time step. According to Theorem \ref{thm:power}, $\bar{y}_t^i$ and $\tilde{y}_t^i$ satisfy normal distribution, which denote as $\bar{y}_t^i \sim \mathcal{N}(\bar{\mu}^i, \bar{\sigma}^i)$ and $\tilde{y}^i_t \sim \mathcal{N}(\tilde{\mu}^i, \tilde{\sigma}^i)$ respectively, then we have $\bar{\mu}^i=\tilde{\mu}^i$, $\bar{\sigma}^i<\tilde{\sigma}^i$. Furthermore for $\forall \xi >0$:
\begin{eqnarray*}
\Pr{(|\bar{y}^i_t-\bar{\mu}^i|<\xi)}=\Pr{(|\frac{\bar{y}^i_t-\bar{\mu}^i}{\bar{\sigma}^i}|<\frac{\xi}{\bar{\sigma}^i})}=2\Phi(\frac{\xi}{\bar{\sigma}^i})-1
\end{eqnarray*}
Similarly, we have:
 $\Pr{(|\tilde{y}^i_t-\tilde{\mu}^i|<\xi)}=2\Phi(\frac{\xi}{\tilde{\sigma}^i})-1$,
where $\Phi()$ represents the probability meets the standard norm distribution. Since $\bar{\sigma}^i<\tilde{\sigma}^i$, then $\Phi(\frac{\xi}{\bar{\sigma}^i})>\Phi(\frac{\xi}{\tilde{\sigma}^i})$, further we can get that $\Pr{(|\bar{y}^i_t-\bar{\mu}^i|<\xi)}>\Pr{(|\tilde{y}^i_t-\tilde{\mu}^i|<\xi)}$.
\end{proof}

\begin{remark}
The corollary ensures the power estimation of our scheme using multi-state setting can attain a more smaller $\mathsf{MAE}$ (see section~\ref{sec:experiments}.Perform metrics) with a higher probability on average than single state-based schemes.
\end{remark}

\section{The Multi-State Dual CNN Model}\label{sec:model}
The discussion above justifies the benefit of the multi-state setting in MS-NILM over the single-state setting. In this section, we propose a novel model for MS-NILM called {\em Multi-State Dual CNN} (MSDC). We begin with an overview of the architecture, followed by the elaboration of its mechanisms. Code and data used for MSDC can be found from our link\footnote{https://github.com/sub-paper/MSDC-NILM\label{code}}.

\begin{figure}
	\centering
	\includegraphics[width=0.48\textwidth]{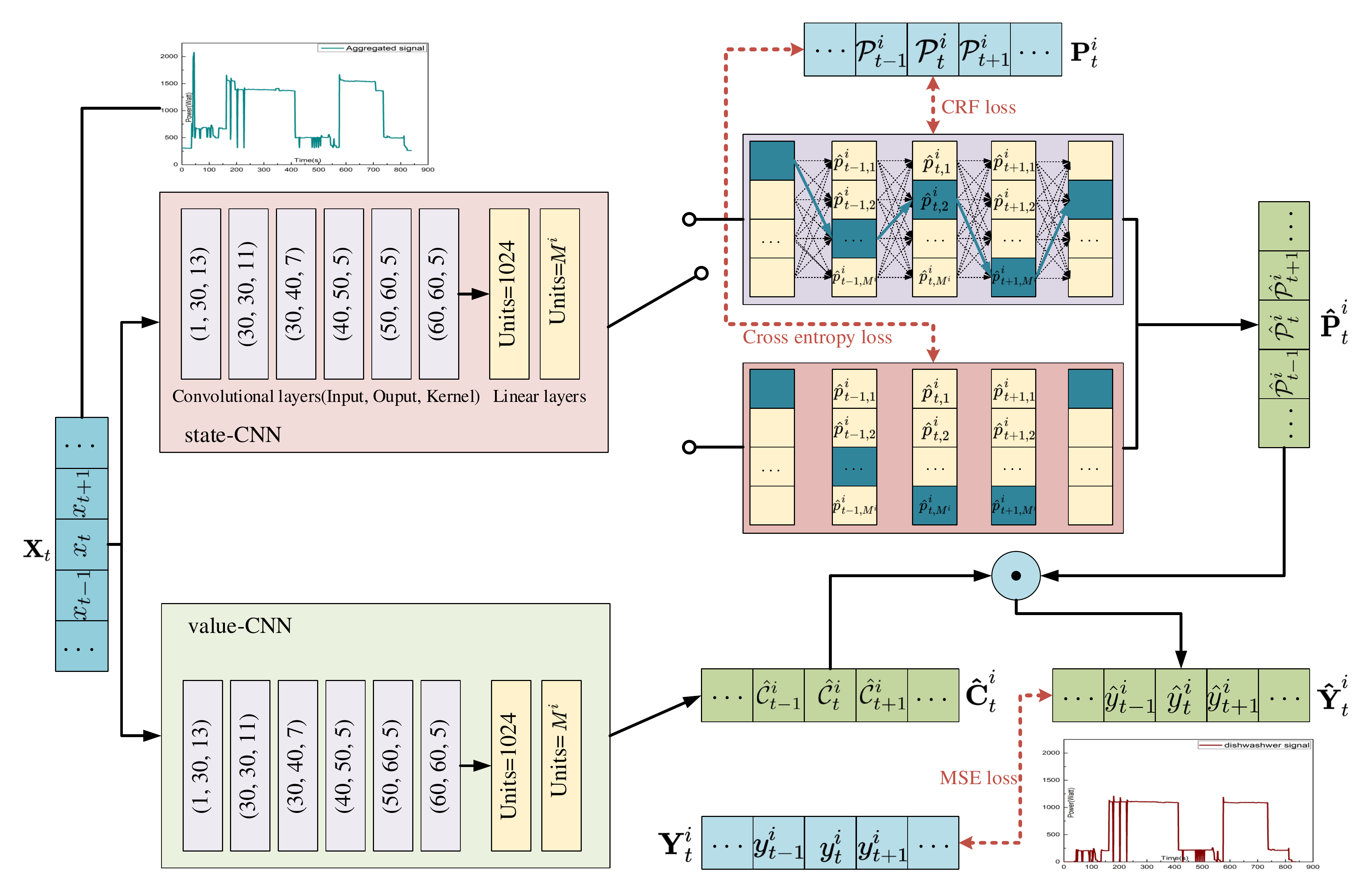}
	\caption{Illustration of our model that features a dual-CNN architecture. Two versions, MSDC and MSDC-CRF, can be obtained by switching between the cross-entropy loss and the CRF loss.}
	\label{fig:CNN-CRF}
	\vspace{-10pt}
\end{figure}

\subsection{Overview}
As Fig.~\ref{fig:CNN-CRF} shows, we train a model for each appliance $i\in [N]$. The dataset consists of the aggregated power signal $\Xv$ as input and  $\Yv^i$ of appliance $i\in [N]$ as output. From $\Yv^i$, an exogenous procedure pre-extracts a fixed set of $M^i$ power states as well as the sequence of power states $s_1^i, s_2^i, \ldots , s_T^i$ of appliance $i$.  Obtaining this information is possible as (1) the power consumption of
common household appliances exhibit clear power-state patterns: a power state appears as a consecutive sequence of relatively stable time steps and a transition takes place when a drastic shift occurs at a time step followed by another power state, whose signal is visibly separated from the previous state; and (2) the states can be conveniently extracted using statistical methods, e.g., a clustering algorithm that groups time steps together if they have similar power consumption levels.

 To address the MS-NILM problem, our model produces the probability distribution $\Pmc_t^i$ over the (pre-extracted) set of power states and the power consumption vector $\Cmc_t^i$ as its output. This is done using a {\em dual-CNN} architecture consisting of a {\em state-CNN} (for $\Pmc_t^i$) and a {\em value-CNN} (for $\Cmc_t^i$).
%
Training the model requires a two-part loss function: First a prediction loss is defined over the disparity between the predicted power consumption signal $\Pmc_t^i\cdot \Cmc_t^i$ and label $\Yv^i$, and then a state-based loss measuring inconsistency between the predicted power state sequence and the pre-extracted sequence.

A simple way to define the state-based loss is simply compare the probability distribution $\Pmc_t^i$ against the pre-extracted power state $q_t^i$. This, however, omits any pattern of state transitions of the appliance. As each appliance is seen as a finite state machine, the shifts between consecutive states in the power state sequence must follow certain patterns that correlate power states. To capture these patterns, we proposed a novel technique that employs {\em conditional random fields} (CRF) \cite{LaffertyMP01}, a discriminative model for exploiting the correlations between consecutive elements of sequential data.
Specifically, we use CRF as an regularization for the state-CNN to explicitly compute the contextual correlations between states. The model is illustrated in Fig.~\ref{fig:CNN-CRF}.  Details of both pre-extracting power states and CRF follow shortly in this section. 

\subsection{Model Description}\label{sec:model-des}

\noindent{\bf Pre-extracting states.}
The procedure that pre-extracts power state sequence from $\Yv^i$ naturally plays a significant role in determining the performance of the model, yet evidence has shown that this step can be easily accomplished for common household appliances whose power states are clearly distinguishable. For example, the {\em mean shift clustering} algorithm \cite{fukunaga1975estimation,cheng1995mean} that features assigning clusters to data without manually pre-defining the number of clusters, could be a useful tool. The cluster centers are determined in an iterative process through computing the mean of the samples in a certain region
It allows us to extract state labels for each appliance by inputting the appliance's power readings: $\Sv^i$ = MeanShift$(\Yv^i)$, where $s_t^i \in [M^i]$ and $M^i$ is automatically determined according to the appliance's power readings $(\Yv^i)$.

\noindent{\bf Model structure.} Aiming at the original power sequence, previous studies employ the sliding window method to overcome the long-sequence issue: separating the original data into a series of short-sequences/windows of equal length. We use a variant of the {\em sliding window} method to overcome the long-sequence issue, which predicts a subsequence centered at the midpoint of the input window. An input window is obtained by dividing the aggregated power signal into several $w$-length segments (can be overlapping). The two CNNs share the same input window, denoted by $\Xv_{t,w}=(x_{t-\lfloor\frac{w}{2}\rfloor}, \ldots, x_{t+\lceil\frac{w}{2}\rceil-1})$. We use $f_{\state}^i$ and $f_{\power}^i$ to represent two CNNs, respectively. The output of our model is a shorter window centered at $t$, which represents a sequence of predicted power signals of an individual appliance. We formally write the output window as
${\hat{\Yv}}_{t,q}^i=(\hat{y}_{t-\lfloor\frac{q}{2}\rfloor}^i, \ldots, \hat{y}_{t+\lceil\frac{q}{2}\rceil-1}^i )$, where $q<w$.

We denote the {\em{state-CNN}} by $f_{\state}^i\colon\mathbb{R}^w_+ \rightarrow [0,1]^{M^i\times q}$, i.e., it outputs a window of predicted state distributions:

\begin{equation}
\hat{\Pv}^i_{t,q}=f_{\state}^i(\Xv_{t,w}),
\end{equation}

where $\hat{\Pv}^i_{t,q} = (\hat{\Pmc}_{t-\lfloor\frac{q}{2}\rfloor}^i, \ldots, \hat{\Pmc}_{t+\lceil\frac{q}{2}\rceil-1}^i )$ and $\hat{\Pmc}_{\tau}^i = (\hat{p}_{\tau,1},  \ldots, \hat{p}_{\tau,M^i})$. Analogously, we represent the {\em{value-CNN}} by $f_{\power}^i\colon\mathbb{R}_+^w \rightarrow \mathbb{R}_+^{M^i\times q}$ that predicts power consumption of each state at each time step:
\begin{equation}
\hat{\Cv}^i_{t,q}=f_{\power}^i(\Xv_{t,w}),
\vspace{-3pt}
\end{equation}

where $\hat{\Cv}^i_{t,q} =  (\hat{\Cmc}_{t-\lfloor\frac{q}{2}\rfloor}^i, \ldots, \hat{\Cmc}_{t+\lceil\frac{q}{2}\rceil-1}^i)$, and $\hat{\Cmc}_{\tau}^i = (\hat{c}_{\tau,1},  \ldots, \hat{c}_{\tau,M^i})$.
%
To simplify the exposition, we will omit $w$ ans $s$ in some notations. By Eq.~\eqref{eq:appliance-power}, $\hat{\Yv}_{t,q}^i$ is derived from the element-wise product of the outputs of two CNNs. More formally, we denote the combination of two CNNs by $f^i_{\comb}\colon\mathbb{R}_+^w \rightarrow \mathbb{R}_+^q$ such that
\begin{equation}
\small
\hat{\Yv}_{t,q}^i = f_{\comb}^i(\Xv_{t,w}) = f_{\state}^i(\Xv_{t,w}) \odot f_{\power}^i(\Xv_{t,w}).
\end{equation}
\smallskip
Finally, the state at each time step can subsequently be determined as the one with the maximum probability: $\hat{\Sv}^i_{t,q}=\arg\max(\hat{\Pv}^i_{t,q} )$.

\vspace{2mm}
\noindent{\bf Loss functions.}
The cost function of the model consists of two parts. The first part considers the error in power consumption prediction, which we measure by the mean squared error averaged over all output windows:
\begin{equation}
\small
\vspace{-3pt}
J_{\power}^i= \Ebb_{t \in [T]} \left[ \left(y_t^i-\hat{\Pmc}_t^i \hat{\Cmc}_t^i \right)^2 \right],
\vspace{-3pt}
\end{equation}
where $y_t^i$ represents the true power signal of the $i$th appliance, and $\hat{\Pmc}_t^i, \hat{\Cmc}_t^i$ are outputs of two CNNs as defined above. The second part penalizes the difference between predicted and true states. We propose two options for realizing it. The first option is to simply ignore the dependence in state shifts and thereby we can use the averaged cross-entropy to measure the difference between the predicted and true states:
\begin{equation}\label{eq:loss-state}
\small
J_{\state}^i=-\Ebb_{t \in [T]} \left[\sum\nolimits_{s=1}^{M^i} p_{t,s}^i \log \hat{p}_{t,s}^i\right],
\end{equation}
where the true state distribution $\Pmc_t^i = (p_{t,1}^i,\ldots, p_{t,M^i}^i)$ is estimated from pre-extracted states. The final cost function sums over two parts: $J^i_{\MSDC}=J_{\state}^i + J_{\power}^i$. In the paper, we use the name {\bf MSDC} to represent the model with the loss function defined above. We next give the second option to realize the loss in terms of state prediction.
\subsection{The CRF Regularization}
In order to capture the state transition, we replace the cross entropy loss $J^i_\state$ in Eq.~\eqref{eq:loss-state} with the CRF regularization that explicitly computes correlations between states
Intuitively, Eq.~\eqref{eq:loss-state} implies that the output of the {\em state-CNN} is a sequence of {\em independent} state distributions. The state at each time step can subsequently be determined as the one with the maximum probability. Training the state-CNN can thus be viewed as solving $q$ independent $M^i$-classification problems.
While, the state distributions in the output window are not independent in the face of state transitions. The CRF regularization allows us to capture state transition by train the {\em state-CNN} in a state sequence-centered way, i.e., view a state sequence as the minimum element in calculating the difference between the predicted and true states, rather than independently consider the state for each time step. Since there are $(M^i)^q$ possible state sequences, training the state-CNN thus turns to solving one $(M^i)^q$-classification problem.
The CRF regularization amounts to a maximum likelihood estimation (MLE) in terms of state sequences. The MLE objective is a log-likelihood consisting of two parts: one is called the {\em emission score} that captures the likelihood of each true state; the other is the {\em transition score} that captures the transition probability between each two neighboring states. It can be formally written as follows:
\begin{equation*}
\vspace{-2pt}
\small
J_{\CRF}^i= -\Ebb \left[\sum\nolimits_{t=1}^T \hat{p}_{t,s_t^i}^i + \sum\nolimits_{t=1}^{T-1}\Psi(s_t^i, s_{t+1}^i) \right] + \log{Z},
\vspace{-1pt}
\end{equation*}
where the expectation is taken on all input state sequences.
The sum $\sum_{t=1}^T \hat{p}_{t,s_t^i}^i$ is namely the emission score, $\Psi(s_t^i,s_{t+1}^i)$ is the transition score computed from all predicted state distributions
, and $Z$ denotes the partition function for two scores, i.e., the total score for all input state sequences. We adopt {\bf MSDC-CRF} for the model with CRF regularization, whose loss function is namely $J^i_{\MSDC-\CRF}=J_{\CRF}^i + J_{\power}^i$.
\begin{table*}
\setlength\tabcolsep{2.5pt}
\small
	\vspace{-2mm}
    \begin{center}
    \scalebox{0.9}{
	\begin{tabular}{ l l l l l l l l}
		\toprule
		Scheme&Fridge&Dishwasher&Microwave&Washing machine&Average&Improvement\\
		\midrule
        FHMM&95.96/29.03/49.17&180.35/713.80/168.34&41.56/79.92/27.15&219.22/498.89/213.46&134.27/330.41/114.53&--/--/--\\
        LSTM&39.59/17.72/22.57&25.43/54.89/24.41&30.99/19.48/24.43&13.23/4.52/11.03&26.31/24.15/20.61&--/--/--\\
        S2P&37.66/15.47/17.66&19.86/15.81/15.24&27.19/21.37/19.33&13.27/4.58/10.84&24.50/14.31/15.78&--/--/--\\
        BERT4NILM&\textbf{28.98}/31.49/17.66&23.31/88.76/23.05&\textbf{17.53}/76.80/15.49&16.68/8.75/12.03&21.63/51.90/17.06&--/--/--\\
        SGN&33.61/19.23/18.45&17.85/38.01/10.33&22.13/49.84/16.35&\textbf{12.97}/2.30/\textbf{10.05}&21.64/27.35/13.80&0.00\%/0.00\%/0.00\%\\
        MSDC&31.78/7.78/\textbf{15.59}&13.03/\textbf{18.44}/\textbf{6.57}&20.36/26.7413.51&13.63/4.34/11.79&19.70/14.3311.87&8.96\%/47.61\%/13.99\%\\
        MSDC-CRF&30.64/\textbf{5.15}/16.08&\textbf{12.81}/21.87/6.75&\textbf{20.35}/23.81/\textbf{12.94}&13.47/\textbf{2.28}/10.65&\textbf{19.32}/\textbf{13.28}/\textbf{11.61}&\textbf{10.72\%}/\textbf{51.44\%}/\textbf{15.87\%}\\
        \bottomrule
	\end{tabular}}

    \smallskip
    \scalebox{0.9}{
	\begin{tabular}{ l l l l l l l l l}
		\toprule
		Scheme&Kettle&Fridge&Dishwasher&Microwave&Washing machine&Average&Improvement\\
		\midrule
        S2P&18.95/26.78/9.25&28.04/31.79/13.71	&26.01/27.82/16.41	&12.35/14.23/6.25	&10.85/\textbf{4.35}/9.54	&19.24/20.99/11.03&--/--/--\\
        BERT4NILM&\textbf{5.02}/\textbf{10.94}/\textbf{3.46}&31.22/35.04/19.91	&34.25/81.11/33.53	&\textbf{6.57}/99.80/6.58	&10.31/41.36/9.45	&17.47/53.65/14.58&--/--/--\\
        SGN&11.81/25.28/10.38	&21.19/14.28/9.07	&\textbf{15.05}/17.47/\textbf{11.12}	&\textbf{7.06}/80.31/6.77	&13.27/26.79/11.85	&13.68/32.83/9.94&0.00\%/0.00\%/0.00\%\\
        MSDC&11.14/17.88/7.63	&\textbf{16.34}/\textbf{6.49}/\textbf{4.42}	&19.21/\textbf{15.75}/13.60	&9.92/\textbf{45.71}/\textbf{6.10}	&\textbf{7.46}/22.50/\textbf{4.93}	&\textbf{12.81}/\textbf{21.67}/\textbf{7.34}	&\textbf{6.36\%}/\textbf{33.99\%}/\textbf{22.13\%}\\
        \bottomrule
	\end{tabular}}
    \end{center}
	\vspace{-3mm}
	\caption{Results for MAE $/$SAE $/$SAE$_{\delta}$ in REDD(top) and UK-DALE(bottom). The bold numbers indicate the best results.}\label{table:general}
\end{table*}

\section{Experiments}\label{sec:experiments}
We evaluate our model on the two most commonly used NILM benchmark datasets, REDD and UK-DALE. We seek to answer the two questions via the experiments: (1) ({\em generalization capability}) Using a set of training data sampled from one household, can our model generalize to other (unseen) households? (2)  ({\em predictive ability})
Can our model accurately predict the power consumption of individual appliances in the household where the training data is sampled?

\subsection{Setup}
\noindent\textbf{Datasets.}
REDD consists of power readings of $6$ households, where the aggregated power and individual appliances' power are recorded every $1$ and $3$ seconds, resp. We preprocess the dataset to remove unusable data and choose four appliances (microwave, washing machine, dishwasher, and fridge) in houses $1$, $2$, and $3$ for evaluation. UK-DALE accommodates power readings of $6$ UK households, which are recorded during a period of $26$ months (from $11/2012$ to $1/2015$). Both aggregated and per-appliance power readings are measured every $6$ seconds. After excluding the unusable data, we select data of 5 appliances (kettle, microwave, washing machine, dishwasher, and fridge) in house $1$ and $2$ for use. We choose these five types of appliances for three reasons: (1) Generalization capability of our model is the most critical factor we aim to evaluate. For this, the training and testing data should be taken from different houses. This means that the same type of appliances must be installed in more than one house. We thus eliminate appliances such as air conditioners and fans, which are only installed in a single house in the datasets. We further exclude appliances whose power consumption data are almost all zero values caused by equipment failure or measuring anomalies. (2) As stated in \cite{kelly2015neural,zhang2018sequence}, these five types of appliances consume a significant portion of household energy. Furthermore, they represent a wide range of possible `power spectra' from the simple on/off-states of a kettle to the complex multiple states of a dishwasher. (3) The datasets and the appliances are chosen by all previous work where baselines were introduced, providing the grounding for fair comparisons.

\noindent\textbf{Baselines.} We compare our model against five baselines: 
{\bf (1)} {\bf FHMM} \cite{kim2011unsupervised}. The (classical) HMM-based model in which each state corresponds to an abstract state that absorbs several states of an appliance.
{\bf (2)} {\bf LSTM} \cite{kelly2015neural}. The first RNN architecture for NILM, which utilizes the long short-term memory (LSTM) network.
{\bf (3)} {\bf S2P} \cite{zhang2018sequence}. A single-state CNN-based model that uses the sequence-to-point strategy, i.e., output only the middle point instead of the entire window.
{\bf (4)} {\bf BERT4NILM} \cite{yue2020bert4nilm}. A bidirectional transformer model with 2 transformer layers and 2 attention heads within each layer.
{\bf (5)} {\bf SGN} \cite{shin2019subtask}. A  state-of-the-art model with a dual-CNN architecture that can capture the appliance's on-off states. SGN achieves state-of-the-art performance on our datasets.

\noindent\textbf{Performance metrics.}
We adopt the following three metrics as performance indicators. {\bf(1)} {\em Mean Absolute Error} ($\mathsf{MAE}$). $\mathsf{MAE}$(Watt) is a general metric to measure the estimation error at each time point. It is formally computed by:
$\mathsf{MAE}^i=\frac{1}{T}\sum_{t=1}^T |\hat{y}^i-y^i|.$
{\bf(2)} {\em Normalized Signal Aggregate Error ($\mathsf{SAE}$)}. $\mathsf{SAE}$(\%) measures the total estimation error in total test time:
$\mathsf{SAE}^i=|\hat{r}^i-r^i| / r^i,$
where $\hat{r}^i=\sum_{i=1}^T \hat{y}^i$ and $r^i=\sum_{i=1}^T y^i$. This metric is also adopted in S2S \cite{zhang2018sequence}.
{\bf(3)} {\em A Variant of $\mathsf{SAE}$} ($\mathsf{SAE}_\delta$). $\mathsf{SAE}_\delta$(Watt) is a variant of $\mathsf{SAE}$ that measures the average total error in a sub-period of the total time: $\mathsf{SAE}_\delta^i=(\sum_{k=1}^{T_\delta}\frac{1}{N_\delta}|\hat{r}_k^i-r_k^i|) \big/ T_\delta$.
Here, $\delta$ represents a physical time period. The total measured time are split to $T_\delta$ time periods, each of an equal length of $N_\delta$ time steps. Following  \cite{shin2019subtask}, we set $\delta=$ 1 hour and $N_\delta=1200$. The average predicted value and true value in the $k$th period are captured by $\hat{r}_k^i=\sum_{t=1}^{N_\delta}\hat{y}^i_{k+t}$ and $r_k^i=\sum_{t=1}^{N_\delta} y^i_{k+t}$, respectively.

\vspace{2mm}
\noindent\textbf{Parameter settings.}
We train one model per appliance. Each of two CNNs consists of $6$ convolutional layers plus $2$ fully connected layers. The CNNs are implemented by Python and Pytorch $1.4.0$+cuda $10$+cudnn$7$, and trained on machines with GTX 1070 Ti (8G) + Ryzen 7 1700 (16 cores). The CRF is implemented by invoking a Pytorch package.
The size of the input/output window is set as $w=400/s=64$ for REDD and $w=200/s=32$ for UK-DALE. Following \cite{zhang2018sequence}, for both datasets, we normalize the power readings beforehand through subtracting the mean values and dividing them by the standard deviations. We use the \emph{mean shift clustering algorithm} to pre-extract appliances' power states and publicize corresponding label data (see our code link~\textsuperscript{\ref{code}}). In each experiment, results for the report are averaged over $20$ independent runs.

\begin{figure}[htb]
\vspace{-5pt}
	\centering
	{ \includegraphics[width=.15\textwidth]{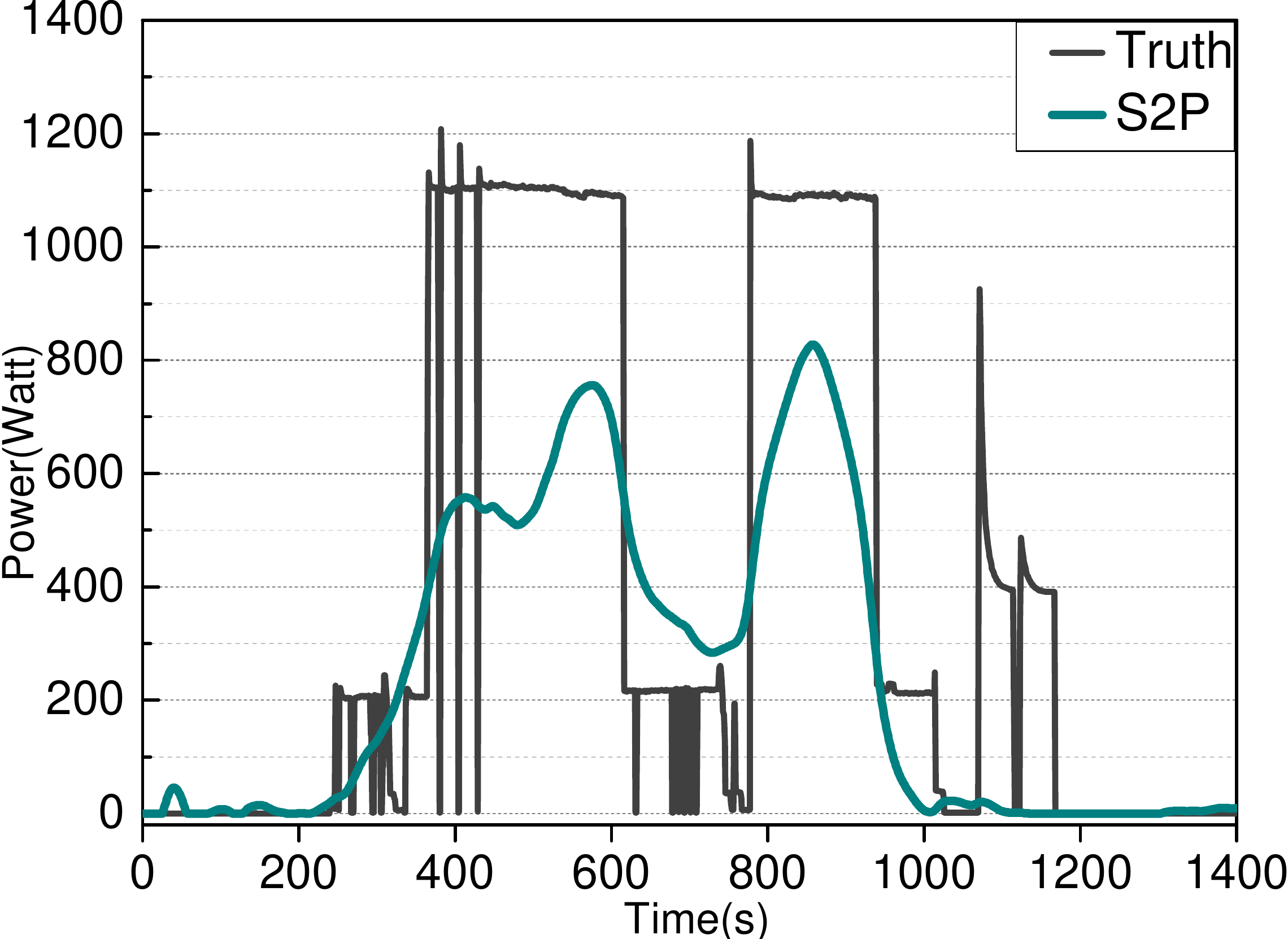} \vspace{-3pt}
        \label{S2S}}
	~~
    \hspace{-0.2in}
	{\includegraphics[width=.15\textwidth]{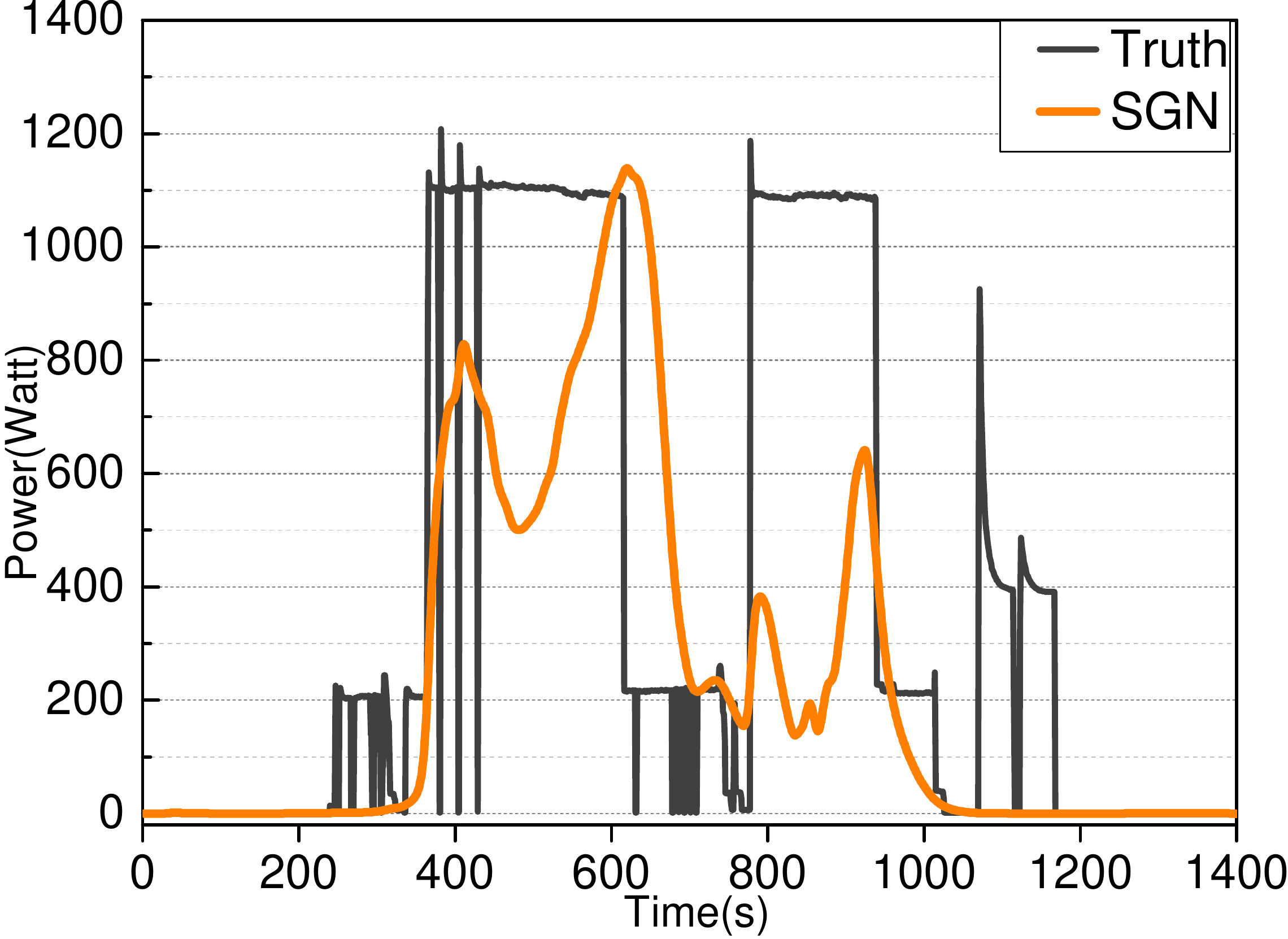} \vspace{-3pt}
    \label{SGN}}
	~~
    \hspace{-0.2in}
	{ \includegraphics[width=.15\textwidth]{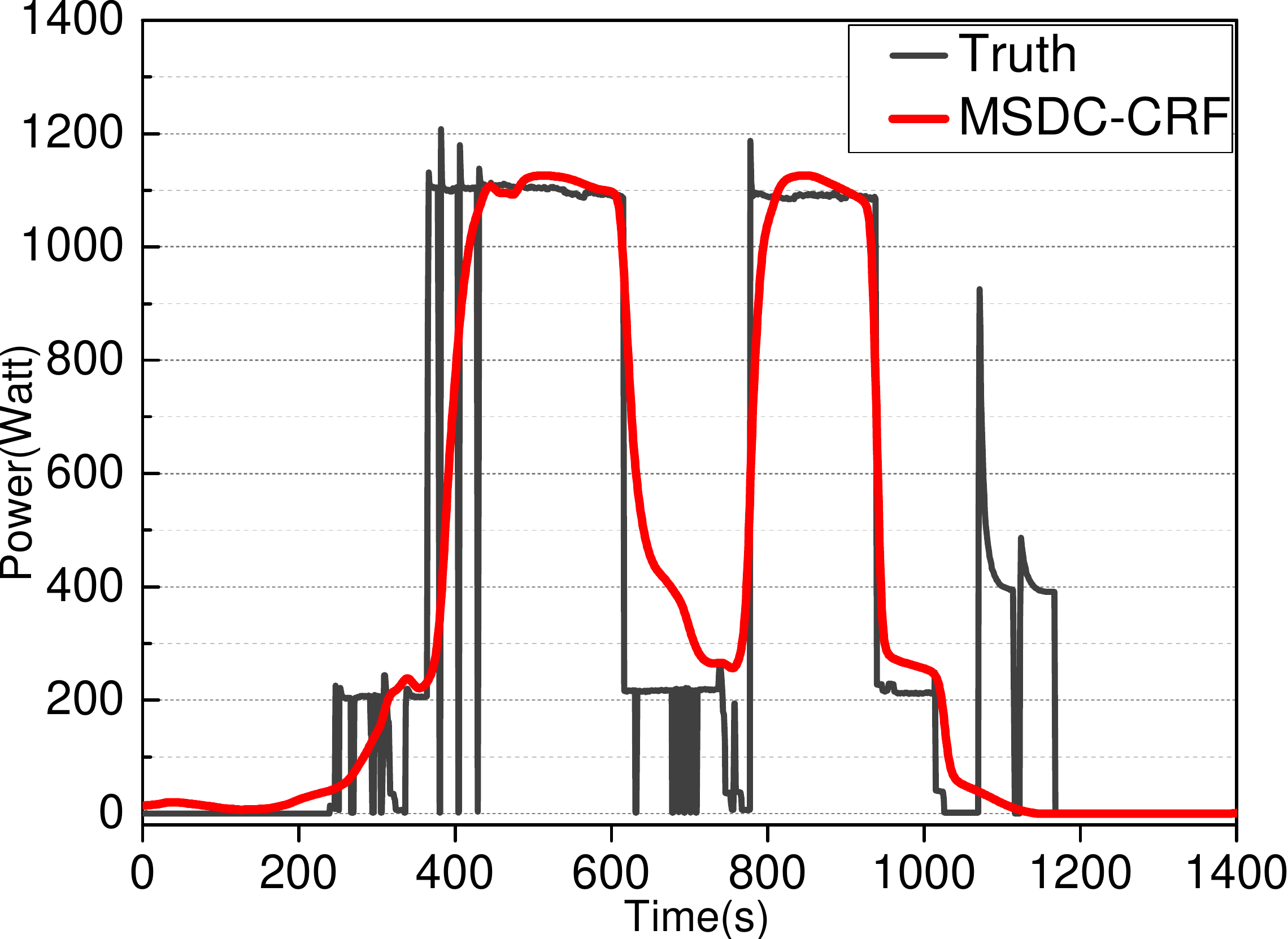} \vspace{-3pt}
        \label{FA}}
	~~
    \caption{Predicted power curves of the dishwasher in house 1 of REDD. Training data is from houses 2, 3. }
    \label{fig:schemes}
    \vspace{-10pt}
\end{figure}

\begin{table}
\renewcommand\arraystretch{0.5}
\small
\setlength\tabcolsep{1.7pt}
    \begin{center}
    \scalebox{0.85}{
	\begin{tabular}{ l l r r r r r}
		\toprule
		Model & Metric & house1 & house2 & house3 & Average & Improvement\\
		\midrule
        &$\mathsf{MAE}$ &8.18  &4.18 &12.83& 8.40 & --\\
        S2P &$\mathsf{SAE}$ &23.87 &29.61 &74.35 & 42.61& --\\
        &$\mathsf{SAE}_\delta$  &5.59&2.90&12.67&7.05& --\\
        \midrule
        &$\mathsf{MAE}$&7.46&1.52&10.27&6.42&0.00\%\\
        SGN&$\mathsf{SAE}$ &15.46&\textbf{5.18}&89.90&36.85&0.00\%\\
        &$\mathsf{SAE}_\delta$ &5.36&0.70&10.38&5.48&0.00\%\\
        \midrule
        &$\mathsf{MAE}$& \textbf{4.96} &\textbf{1.09} &\textbf{7.87} &\textbf{4.64}&\textbf{27.73}\%\\
        MSDC&$\mathsf{SAE}$ &8.49  &7.87 &\textbf{59.27} &\textbf{25.21}&\textbf{31.59}\%\\
        &$\mathsf{SAE}_\delta$ &2.81  &0.56 &\textbf{7.31} &\textbf{3.56}&\textbf{35.04}\%\\
        \bottomrule
	\end{tabular}}
	\smallskip
	
	\scalebox{0.85}{
	\begin{tabular}{l l r r r r}
	    \toprule
		Model & Metric & house1 & house2 & Average & Improvement\\
		\midrule
        &$\mathsf{MAE}$&12.50&4.99&8.75&--\\
        S2P &$\mathsf{SAE}$ &34.92&1.51&18.72&--\\
        &$\mathsf{SAE}_\delta$ &9.16&2.36&5.76&--\\
        \midrule
        &$\mathsf{MAE}$&6.18 &4.04 &5.11&0.00\%\\
        SGN&$\mathsf{SAE}$ &12.52&3.29&7.91&0.00\%\\
        &$\mathsf{SAE}_\delta$ &5.30& 2.86 &4.08&0.00\%\\
        \midrule
        &$\mathsf{MAE}$ &\textbf{3.24}&\textbf{3.51} &\textbf{3.38}&\textbf{33.86\%}\\
        MSDC&$\mathsf{SAE}$ &\textbf{3.49}&\textbf{3.51}&\textbf{3.38}&\textbf{33.86\%}\\
        &$\mathsf{SAE}_\delta$ &\textbf{2.45} &\textbf{2.33} &\textbf{2.39}&\textbf{41.42\%}\\
        \bottomrule
	\end{tabular}}
    \end{center}
    \vspace{-2mm}
    \caption{Results for the dishwasher in REDD (top) and  UK-DALE (bottom). Best results are highlighted in bold.}
    \label{table:houses}
    \vspace{-12pt}
\end{table}

\subsection{Generalization Capability}
To investigate the generalization capability of our model, we take training and testing data from different houses: for REDD, house $2$ and $3$ are for training and house $1$ for testing; for UK-DALE, the setting turns to house $1$ for training and house $2$ for testing.

The top of Tab.~\ref{table:general} shows the comparisons of our MSDC and MSDC-CRF against baselines on REDD. All the deep learning-based models perform better than FHMM and our models perform the best on almost all appliances. MSDC achieves an average improvement for $\mathsf{MAE}$, $\mathsf{SAE}$, and $\mathsf{SAE}_\delta$ by $8.96\%$, $47.61\%$, and $13.99\%$, respectively, compared to the best baseline. Further, MSDC-CRF demonstrates even higher improvements on all metrics ($10.72\%$, $51.44\%$, and $15.87\%$ on average). This demonstrates the advantage of using CRF to capture state transition.

We further compare these baselines that attain better average performance. From the results on UK-DALE (the bottom of Tab~\ref{table:general}), we can see that our MSDC performs best on average, achieving an improvement of $6.36\%$-$33.99\%$ on the standard metrics.
From Tab~\ref{table:general} we may find that BERT4NILM can achieve better results on some metrics, it, however, cannot ensure to produce robust and consistent performance. For example, BERT4NILM obtains a much higher $\mathsf{SAE}$ on most appliances compared to SGN. On the other hand, BERT4NILM is a transformer that is a very costly option, which requires more time for model training.

We then visualize the results to intuitively display the difference between the predicted results and the ground truth data. Figure \ref{fig:schemes} depicts the predicted power curves for one working cycle of the dishwasher. For a clear demonstration, we only show results of S2P, SGN, and our MSDC-CRF. Among all models involved, MSDC-CRF attains the smallest prediction error and predicts on/off states with the highest accuracy. Moreover, the power levels of all ``on'' states predicted by our scheme are much more consistent with the truth. Although the average power level for each state in training houses is not the same as in the testing house, MSDC-CRF achieves the highest accuracy, indicating superior generalization capability of our model. However, no model can well predict the power level in the last time period of the cycle (from $1000$s to $1200$s in the figure). The reason may be that the dishwashers in two houses belong to different models, and this time period corresponds to a working state which only exists in the dishwasher of house 1 (test data), but not in the dishwashers in house 2 and 3 (training data). Considering this mismatching issue, we next move our focus to the predictive ability of our model assuming two appliances match in states.

\subsection{Predictive Ability}
We evaluate the prediction ability of our model. To make appliances in training and testing matching in states, for both datasets we take the training and testing data from the same house. We set the ratio of the data for training, validation, and testing as $7$:$1$:$2$. Several previous work \cite{kolter2010energy,elhamifar2015energy,he2019efficient} also adopted a similar setting.
\begin{figure}
	\centering
	{ \includegraphics[width=.15\textwidth]{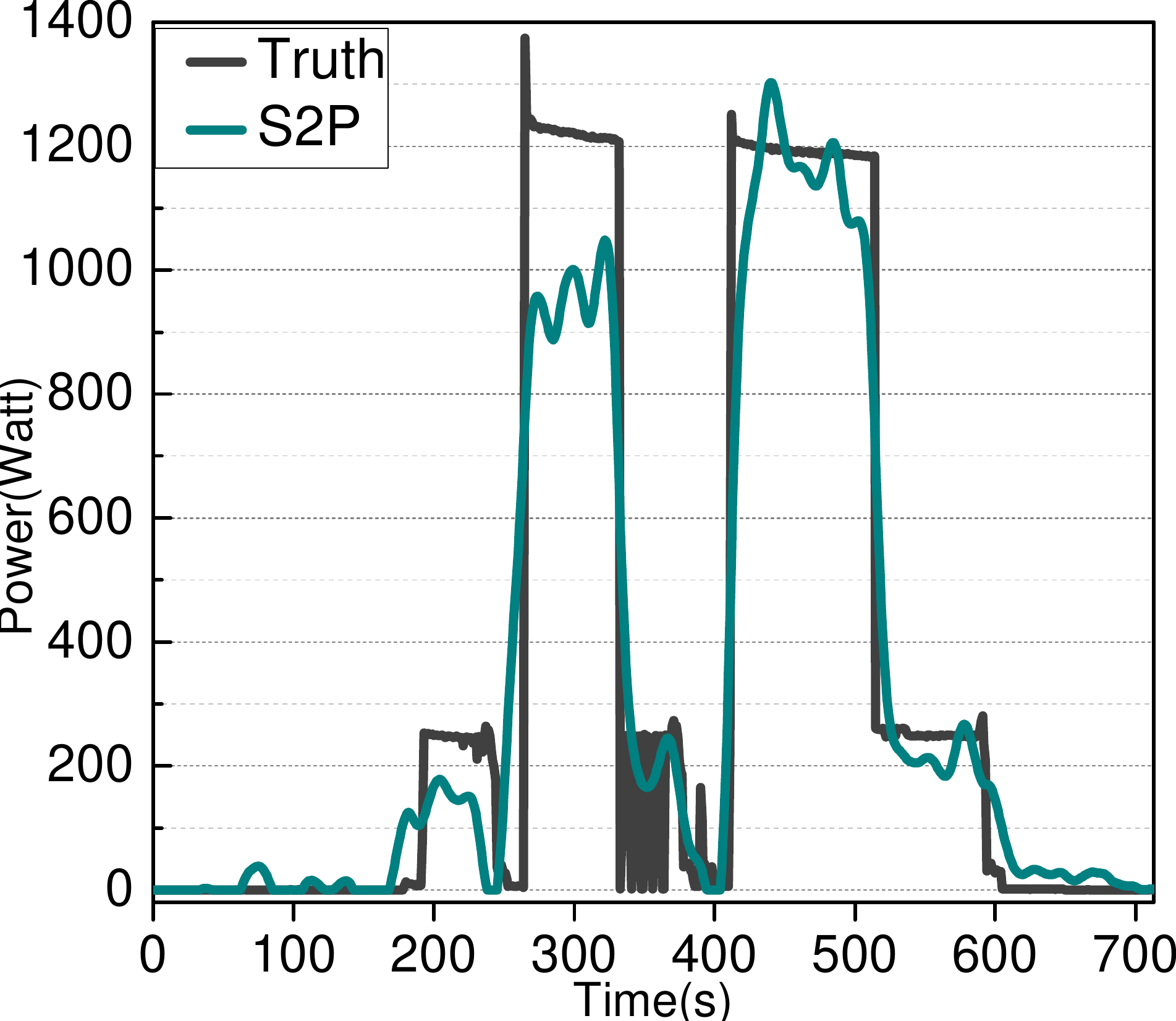} \vspace{-3pt}
        \label{house2-no}}
	~~
    \hspace{-0.18in}
	{\includegraphics[width=.15\textwidth]{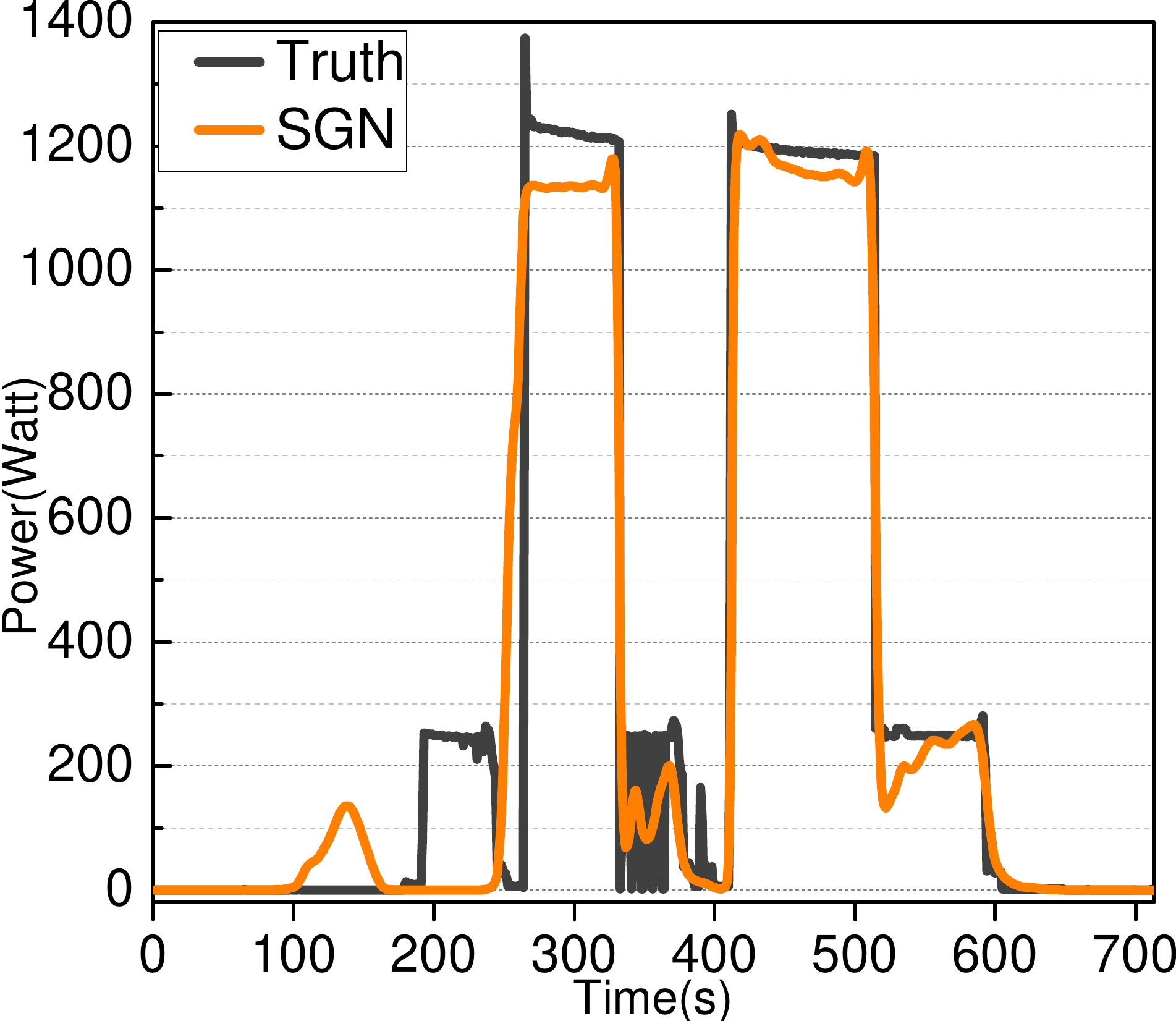}  \vspace{-3pt}
    \label{house2-on}}
	~~
    \hspace{-0.2in}
	{ \includegraphics[width=.15\textwidth]{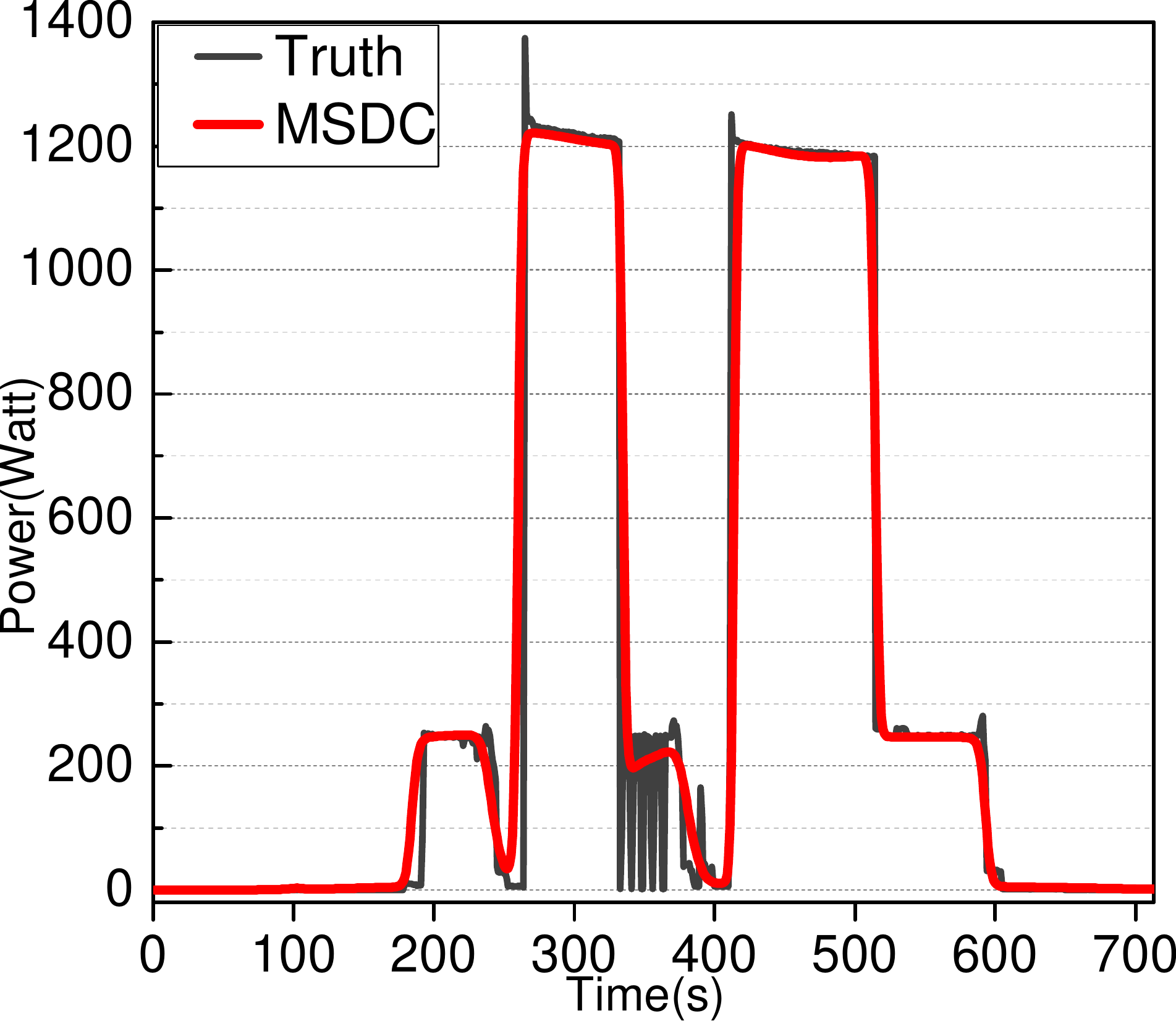} \vspace{-3pt}
        \label{house2-multi}}
	~~
    \caption{Predicted power curves of the dish-washer in house 2 of REDD. Training data is from the same house.} 
    \label{fig:houses}
\end{figure}
Due to space limitation, we only report the results of S2P, SGN, and our MSDC on the dishwasher.
Tab.~\ref{table:houses} records the results of dishwasher for two datasets, which reveal that our model achieves significant performance enhancements in all houses. Specifically, the average improvements on $\mathsf{MAE}$, $\mathsf{SAE}$, and $\mathsf{SAE}_\delta$ are up to $27.73\%$, $31.59\%$, and $35.04\%$, respectively in REDD dataset; $33.86\%$, $33.86\%$, and $41.42\%$, respectively in UK-DALE dataset.
We owe the improved performances to the fine-grained state structure used in our model, in contrast to two single-state baselines.

Fig.~\ref{fig:houses} depicts the predicted power curves for one working cycle of the dishwasher in house $2$ in REDD. The predicted power signal of our model demonstrates a closer resemblance to the true signal compared to baselines.  The signal output by our model is also smoother whereas baseline models output many nonzero values when the dishwasher is actually ``off''.  Moreover, the baseline results exhibit many abnormal fluctuations or peaks, which are not present in the predicted output of our model.

\section{Conclusions and Future work}
In this paper, we extend NILM task to the multi-state setting and formalize the {\em{multi-state-NILM}} (MS-NILM) problem. The problem seeks a model that predict per-appliance power signal through capturing the feature of multiple power states of appliances. We theoretically justify the advantage of the multi-state setting on reducing the sample variance of training data. We then proposed a dual-CNN-based model called MSDC to address  MS-NILM, which uses two CNNs to predict the state distribution and acquire the power consumption of each state, respectively. To capture state transitions, we further design a novel technique that incorporates CRF into MSDC, leading to a variant model called MSDC-CRF. Experimental results show that our model has the excellent ability in recovering power consumption and good capability to predict for unseen appliances.

As future work, one could attempt to generalize MSDC to more general setups, say, Type-3 appliances which have continuous state space. Here possible ideas include discretizing the state space or parametrizing the state functions.  Another promising future work is to
scale our model to a broader range of BSS problems such as speech separation and recognition where each speech signal corresponds to a specific identifiable hidden state.

\section*{Acknowledgments}
This work is supported by National Natural Science Foundation of China under Grant No.62172040, National Key R$\&$D Program of China under Grant No.2022YFB3103500, and National Natural Science Foundation of China under Grants No.U1836212, No.61872041 No.U20A20176 and No.62072062.

\bibliography{aaai23}
 \end{document}

%% file: commands.tex
\usepackage{bm}
\usepackage{dsfont}

\usepackage{times}

\usepackage{soul}
\usepackage{url}
\usepackage[draft]{hyperref}
\usepackage{graphicx}
\usepackage{amsmath}
\usepackage{mathrsfs}
\usepackage{bbm}
\usepackage{amsthm}
\usepackage{booktabs}
\usepackage{graphicx,subfigure}
\usepackage{xcolor}
\usepackage{amsmath,amssymb,amsfonts,multirow}
\newtheorem{theorem}{Theorem}

\newtheorem{corollary}{Corollary}
\newtheorem{remark}{Remark}
\newtheorem{assumption}{Assumption}
\newtheorem{fact}{Fact}
\newcommand{\Cv}[0]{{{\bf C}}}

\newcommand{\Pv}[0]{{{\bf P}}}

\newcommand{\Sv}[0]{{{\bf S}}}

\newcommand{\Xv}[0]{{{\bf X}}}
\newcommand{\Yv}[0]{{{\bf Y}}}

\newcommand{\Cmc}[0]{{{\mathcal{C}}}}

\newcommand{\Nmc}[0]{{{\mathcal{N}}}}

\newcommand{\Pmc}[0]{{{\mathcal{P}}}}

\newcommand{\Ebb}{\mathbb{E}}
\newcommand{\Rbb}{\mathbb{R}}
\newcommand{\Nbb}{\mathbb{N}}

\newcommand{\state}{\mathrm{state}}

\newcommand{\power}{\mathrm{power}}
\newcommand{\comb}{\mathrm{comb}}
\newcommand{\CRF}{\mathrm{CRF}}
\newcommand{\MSDC}{\mathrm{MSDC}}